\documentclass{article}

\usepackage{microtype}
\usepackage{graphicx}
\usepackage{subfigure}
\usepackage{booktabs} 
\usepackage{mathrsfs}

\usepackage{thm-restate}
\usepackage{listings}

\usepackage{nicefrac}
\usepackage{xcolor}
\usepackage{amsmath,amssymb,amsthm}

\theoremstyle{definition}

\usepackage[framemethod=TikZ]{mdframed}

\usepackage{enumitem}
\setlist[enumerate]{leftmargin=.45cm,topsep=0pt,itemsep=-2pt}
\setlist[itemize]{leftmargin=.45cm,topsep=0pt,itemsep=-2pt}

\usepackage{mathrsfs}

\usepackage{mathtools}
\usepackage{hyperref}
\hypersetup{
    colorlinks=true,
    linkcolor=blue,
    citecolor=cyan,
}

\newcommand{\notimplies}{%
  \mathrel{{\ooalign{\hidewidth$\not\phantom{=}$\hidewidth\cr$\implies$}}}}

\newcommand{\E}{\mathbb{E}}

\usepackage[accepted]{icml2021}

\icmltitlerunning{On a few pitfalls in KL divergence gradient estimation for RL}

\begin{document}

\twocolumn[
\icmltitle{On a few pitfalls in KL divergence gradient estimation for RL}




\begin{icmlauthorlist}
\icmlauthor{Yunhao Tang}{meta1}
\icmlauthor{R\'emi Munos}{meta2}

\end{icmlauthorlist}
\icmlaffiliation{meta1}{Work done while at Meta GenAI}
\icmlaffiliation{meta2}{Meta FAIR}


\icmlkeywords{Machine Learning, ICML}

\vskip 0.3in
]



\printAffiliationsAndNotice{} 

\begin{abstract}
We point out a few pitfalls in implementing gradient estimation for KL divergence in RL training for LLM, as seen in a number of open source projects and papers. The first major pitfall is to differentiate through the KL estimate as loss functions to minimize KL divergence. We show that
such implementations are generally incorrect and do not produce the desired KL gradient. Secondly, we show that some implementations do not account for the sequential nature of the estimation problem and produce a partial gradient at best. We demonstrate the impact of such issues with illustrative tabular and LLM experiments, and show the correct way to implement the KL gradient.
\end{abstract}

\section{Introduction}

KL divergence is a long-standing and predominant regularization measure in reinforcement learning (RL) \citep{ziebart2008maximum,schulman2017equivalence,schulman2015trust,schulman2017proximal,haarnoja2018soft} and recently in its applications to large language models (LLM) \citep{christiano2017deep,ziegler2019fine,ouyang2022training}. The idea is to prevent the learned policy from drifting too much during training (e.g., to preserve sample diversity), by regularizing the learned policy against a reference policy. Concretely, this is often achieved by minimizing the KL divergence between the two policies. 

In this short note, we point out a few common pitfalls in implementing the \emph{loss} for minimizing KL divergence. Concretely, one popular practice is to construct a Monte-Carlo estimate to the KL divergence $\mathbb{KL}$ and use its gradient $\nabla \mathbb{KL}$ computed via auto-differentiation, to minimize the divergence. We will see that this strategy does not produce the right gradient in general. As a result, the targeted KL regularization is not enforced as intended!

We also find that common gradient estimates for LLM applications only account for a partial derivative of the full gradient. The source issue is that, since generating a sequence is auto-regressive, past tokens can impact future tokens in a way that is not captured by some loss implementations. We find that accounting for such dependencies improve the ability to enforce KL regularization.

We observe that such conceptual pitfalls might impact a number of practical implementations in open source RLHF projects, such as the TRL \citep{vonwerra2022trl} and Open Instruct \citep{lambert2024tulu3}. Such implementations have also been described in the pseudocode for a number of seminal work (e.g., GRPO  \citep{shao2024deepseekmath,guo2025deepseek}) as well as subsequent papers that heavily borrow such loss designs, as shown in their pseudocode or open source code  (see, e.g., \citep{liu2025understanding,zhao2025d1}).

\paragraph{What this work is not about.} We do not seek to demonstrate the importance of KL divergence or other divergences in LLM training. Instead, assuming that KL regularization is needed, we demonstrate the impact of different gradient estimates (the correctly implemented vs. incorrect ones).

\section{Estimating KL divergence with samples}

Given two distribution $\pi,\pi_\text{ref}$ over the same sample space $y\in\mathcal{Y}$, we define the KL divergence between the two distributions as
\begin{align*} 
    \mathbb{KL}(\pi,\pi_\text{ref}) \coloneqq \mathbb{E}_{y\sim \pi}\left[\log\frac{\pi(y)}{\pi_\text{ref}(y)}\right].
\end{align*}

In general, computing the above expectation exactly is infeasible. Instead, it is more tractable to build stochastic estimations for the KL divergence. 
The estimation of KL divergence has been long studied in the literature and we summarize a few popular options below.


\subsection{A few popular KL divergence estimates}

Assuming a single sample drawn from $y\sim \pi$, and by definition of the KL divergence, we can define the unbiased vanilla estimate 
\begin{align*}
\widehat{\mathbb{KL}}_\text{vanilla}\coloneqq \log\frac{\pi(y)}{\pi_\text{ref}(y)}.
\end{align*}

 Another estimate, though not formally published \citep{schulman-blogpost}, is quite popular among practitioners as commonly implemented in a number of recent work,
\begin{align*}
    \widehat{\mathbb{KL}}_\text{var-reduced}\coloneqq \log\frac{\pi(y)}{\pi_\text{ref}(y)} + \frac{\pi_\text{ref}(y)}{\pi(y)} - 1.
\end{align*}
To better understand the estimate, note that since the importance ratio $\frac{\pi_\text{ref}(y)}{\pi(y)}$ is mean $1$, the above estimate is unbiased too. In that sense, the zero-mean random variable $- \frac{\pi(y)}{\pi_\text{ref}(y)} + 1$ can be understood as a control variate that generally reduces the variance of the overall estimate \citep{robert1999monte}. 

Finally, 
\citet{schulman-blogpost} also introduced a biased estimate motivated by a Taylor expansion view of KL divergence, which achieves smaller errors when $\pi$ and $\pi_\text{ref}$ are close,
\begin{align}
    \widehat{\mathbb{KL}}_\text{squared}\coloneqq \frac{1}{2}\left(\log\frac{\pi(y)}{\pi_\text{ref}(y)}\right)^2.\label{eq:biased}
\end{align}

\begin{figure}[t]
\centering
\includegraphics[width=.45\textwidth]{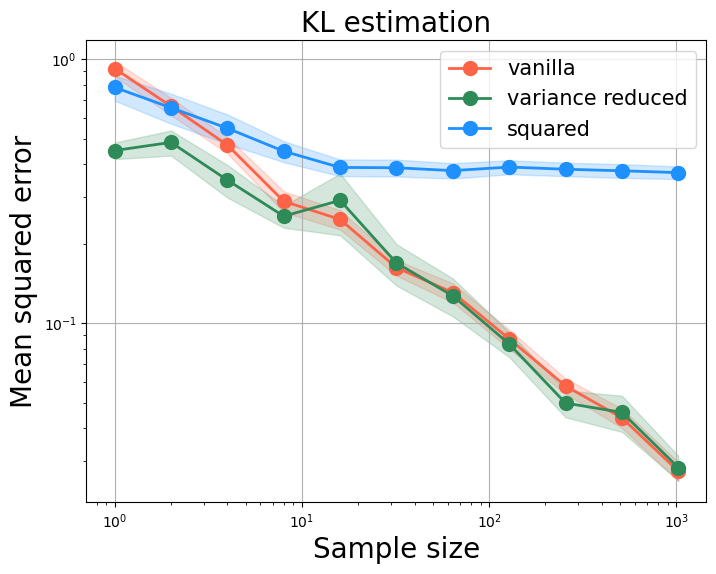}
\caption{\small{Comparing the accuracy of KL divergence estimation. At small sample size, variance reduced and squared estimates have lower mean squared error (MSE) compared to vanilla thanks to lower variance. At large sample size, squared estimate under-performs due to bias.}}
\label{figure:kl}
\end{figure}

\paragraph{Empirical validation.} We validate the aforementioned insights on variance reduction through a tabular experiment in Figure~\ref{figure:kl}. We average over $n$ samples to produce a final estimate, and average over $100$ simulations to produce an estimate of mean squared error (MSE). As shown, at small $n$, the squared estimate has lower MSE than the vanilla estimate though ultimately saturates at a higher MSE due to bias. Overall, the variance reduced estimate has lower MSE than the vanilla estimate, corroborating the observation in \citep{schulman-blogpost}.

\subsection{Minimizing KL divergence by differentiating through KL divergence estimates}

To implement parameter update that minimizes the KL divergence, a common strategy is to define the loss function as the negative of the KL divergence estimate as shown in a number of open source releases \citep{ziegler2019fine,vonwerra2022trl,lambert2024tulu3}. 
The motivation is clear: by applying auto-differentiation (e.g., with PyTorch \citep{paszke1912pytorch} or Jax \citep{jax2018github}) to such a loss, we seem to gradient descend on an unbiased estimate of the KL divergence itself, hence minimizing the objective.

In the following, we will show that differentiating such losses mostly do not produce the desirable gradient. This means that the target KL regularization does not get enforced in the resulting optimization. The following summarizes the key insight of this pitfall.

\begin{figure}[t]
\centering
\includegraphics[width=.45\textwidth]{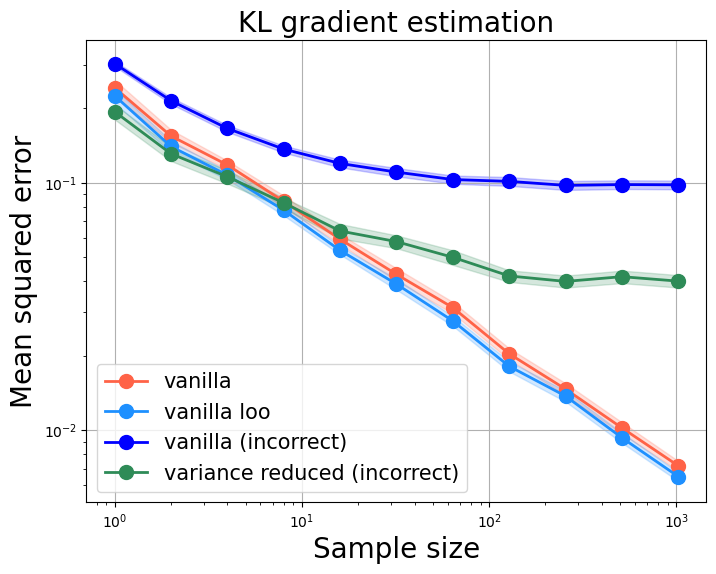}
\caption{\small{Comparing the accuracy of KL gradient estimation. We label biased implementations as \emph{incorrect}. As sample size increases, vanilla estimates obtain lower MSE while estimates resulting from incorrect implementations plateau at large MSE due to bias. The vanilla estimate with leave-one-out control variate obtains marginally smaller variance. In RL for LLM, we find that proper KL gradient estimates make significant difference when the KL divergence is large, e.g., for on-policy distillation.}}
\label{figure:kl-gradient}
\end{figure}

\mdfsetup{%
backgroundcolor=black!10,
roundcorner=10pt}
\begin{mdframed}
\textbf{Key insight.}\\
Differentiating KL estimates as loss functions does not usually produce the right estimate to the KL gradient.
\end{mdframed}

\section{Differentiating unbiased KL estimates do not produce unbiased gradient estimates}

A key technical observation is that the gradient of the aforementioned KL divergence estimates does not produce the desired gradient to KL divergence in general. In more technical terms, by differentiating the KL divergence estimates, we might not obtain an unbiased gradient estimate. 
\begin{align*}
\mathbb{E}\left[\widehat{\mathbb{KL}}\right] = \mathbb{KL} \notimplies 
 \mathbb{E}\left[\nabla \widehat{\mathbb{KL}}\right] = \nabla \mathbb{KL}
\end{align*}
Essentially, because the KL divergence $
\mathbb{KL}(\pi,\pi_\text{ref})$ is an expectation depending on $\pi$ both in terms of the sampling measure and the integrand, its gradient has two components: a score function derivative and a path-wise derivative part \citep{glasserman2004monte}. Simply differentiating the KL estimate $\nabla \mathbb{KL}$ only accounts for the path-wise part, and so in general, does not produce the correct full gradient. We provide a full discussion in Appendix~\ref{appendix:gradient}.

We discuss the few KL estimates introduced above.

\paragraph{Vanilla estimate.} Maybe to the surprise of some practitioners, differentiating the vanilla KL estimate with respect to parameters of $p$, we obtain a gradient evaluated to zero in expectation
\begin{align}
    \mathbb{E}_{y\sim \pi}\left[\nabla \widehat{\mathbb{KL}}_\text{vanilla}\right]=\mathbb{E}_{y\sim \pi}\left[\nabla \log \pi(y)\right]=0.\label{eq:vanilla}
\end{align}
In other words, using such a gradient estimate amounts to adding a zero-mean noise to the aggregate gradient estimate.

\paragraph{Variance reduced estimate.} Differentiating the variance reduced KL  estimate, we obtain a gradient estimate with non-zero expectation. 
\begin{align}
    \mathbb{E}\left[\widehat{\nabla \mathbb{KL}}_\text{var-reduced}\right]= \nabla \mathbb{KL}(\pi_\text{ref},\pi) \neq \nabla \mathbb{KL}(\pi,\pi_\text{ref}).\label{eq:var-reduced}
\end{align}
Very importantly, the positions of the two distributions are reversed compared to the target KL divergence $\mathbb{KL}(\pi_\text{ref},\pi)\neq \mathbb{KL}(\pi,\pi_\text{ref})$, see Appendix~\ref{appendix:derivation} for details. This implies that gradient descent using the above gradient estimate will converge to the same minimizer as the target KL divergence $\pi=\pi_\text{ref}$. However, when combined with reward maximization, the variance reduced estimate converges to a different policy than the target optimal policy.

It is important to understand that the regularization effect of the variance reduced estimate is completely incidental! The regularizer comes from the control variate term of the estimate, whose role is generally considered optional. Indeed, imagine introducing a multiplicative factor $\lambda\in\mathbb{R}$ to the KL estimate as suggested in \citep{schulman-blogpost}
\begin{align*}
    \mathbb{KL}_\lambda\coloneqq \log\frac{\pi(y)}{\pi_\text{ref}(y)} + \lambda\left(\frac{\pi_\text{ref}(y)}{\pi(y)} - 1 \right),
\end{align*}
which is an unbiased estimate of $\mathbb{KL}(\pi, \pi_\text{ref})$ for any value of $\lambda$. Following the same procedure as before, we see that
$\mathbb{E}\left[\nabla  \mathbb{KL}_\lambda\right]=\lambda \mathbb{KL}(\pi_\text{ref},\pi)$ which means that the regularization effect is weighted by $\lambda$, maybe unexpectedly to some practitioners. Indeed, when $\lambda=0$ this recovers differentiating through the vanilla estimate, which has no regularization effect at all.

\begin{figure}[t]
\centering
\includegraphics[width=.45\textwidth]{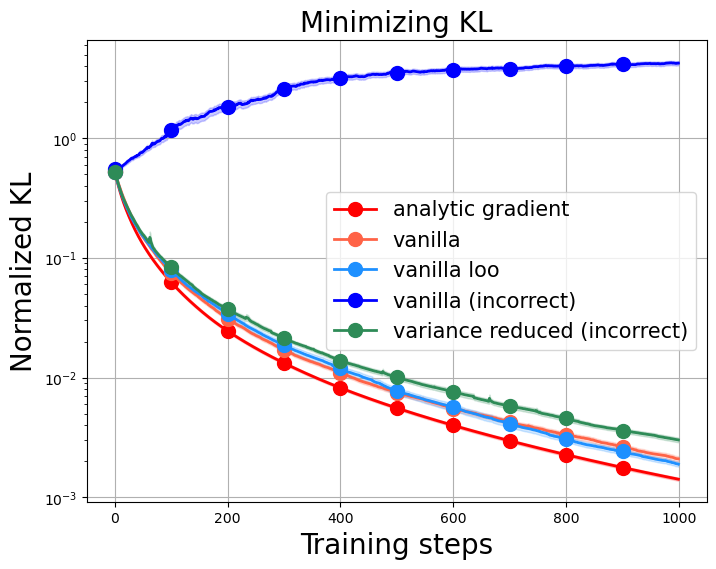}
\caption{\small{Comparing the efficacy of different estimates at minimizing KL divergence in tabular settings. We initialize $\pi$ to be different from $\pi_\text{ref}$ and descend on the KL gradient estimate. Except the incorrectly implemented vanilla estimate, which has zero expected gradient, other estimates generally decrease the KL divergence.}}
\label{figure:kl-minimization}
\end{figure}

\mdfsetup{%
backgroundcolor=black!10,
roundcorner=10pt}
\begin{mdframed}
\textbf{Key insight.}\\
The commonly adopted variance reduced gradient estimate incidentally minimizes reverse-KL divergence.
\end{mdframed}

\paragraph{Squared estimate.} Intriguingly, the squared KL divergence estimate, despite being a biased estimate itself, produces an unbiased KL gradient estimate when differentiated
\begin{align*}
    \mathbb{E}\left[\nabla \widehat{\mathbb{KL}}_\text{squared}\right] = \mathbb{E}\left[\log\frac{\pi(y)}{\pi_\text{ref}(y)}\nabla \log \pi(y)\right] = \nabla \mathbb{KL}(\pi,\pi_\text{ref}).
\end{align*}

For convenience of ensuing discussions, we will also call the resulting gradient estimate the \emph{vanilla gradient estimate}.
\begin{align}
    \log\frac{\pi(y)}{\pi_\text{ref}(y)}\nabla \log \pi(y).\label{eq:vanilla-correct}
\end{align}
Note that differentiating squared estimate is not the only way to implement the vanilla gradient estimate. In practice, it is common to differentiate the surrogate loss  $
    \text{sg}\left(\log\frac{\pi(y)}{\pi_\text{ref}(y)}\right) \log \pi(y)$ to obtain the same gradient estimate, where $\text{sg}$ denotes the stop-gradient operation.

In general, when we build an estimate with $n$ samples $(y_i)_{i=1}^n$, we can construct the Monte-Carlo average $
    \frac{1}{n}\sum_{i=1}^n \log\frac{\pi(y_i)}{\pi_\text{ref}(y_i)}\nabla \log \pi(y_i)$ as the final vanilla estimate. We can also subtract a baseline $v_i$ for variance reduction 
    \begin{align*}
        \frac{1}{n}\sum_{i=1}^n \left(\log\frac{\pi(y_i)}{\pi_\text{ref}(y_i)} - v_i\right)\nabla \log \pi(y_i).
    \end{align*}
A near-optimal baseline is the true KL divergence $v^*=\mathbb{KL}(\pi,\pi_\text{ref})$ itself, which requires summing over the full sample space.
An alternative is the leave-one-out control variate $v_i = \frac{1}{n-1}\sum_{j\neq i} \log\frac{\pi(y_j)}{\pi_\text{ref}(y_j)}$, which approximates $v^*$ while independent of $y_i$ so as not to introduce bias \citep{mnih2016variational,kool2019buy}.

\paragraph{Analytic gradient.} Finally, we note that when the sample space $\mathcal{Y}$ is tractable, it is possible to compute the gradient analytically, via the following formula
\begin{align}
    \sum_{y\in\mathcal{Y}} \pi(y) \log \frac{\pi(y)}{\pi_\text{ref}(y)}\nabla \log \pi(y)\label{eq:analytic}
\end{align}
which is an integrated version of the vanilla gradient estimate and  enjoys strictly lower variance. In practice, this involves additional trade-off between computation and statistical accuracy, which we do not explore further in this work. Whenever computing the analytic gradient is feasible, we use it as the gold standard.

\begin{figure}[t]
\centering
\includegraphics[width=.45\textwidth]{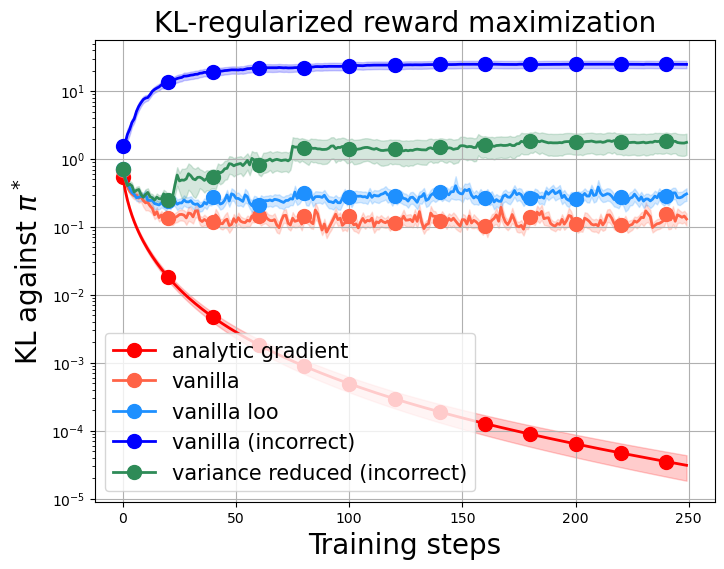}
\caption{\small{Measuring the KL divergence between $\pi$ and optimal policy $\pi^*$ for KL-regularized maximization problems. We initialize $\pi=\pi_\text{ref}$ and follow the gradient estimates that combine the reward and KL components. While vanilla estimates generally decrease $\mathbb{KL}(\pi,\pi^*)$ converging to a value that depends on the sample size, the incorrectly implemented vanilla and variance reduced estimates do not.}}
\label{figure:regularized-kl}
\end{figure}

\subsection{Tabular experiments}

We validate the bias of KL gradient estimates and its effect on optimization in a set of tabular experiments. We use \emph{vanilla (incorrect)} to denote the incorrect differentiation through the vanilla KL estimate to produce a gradient, and \emph{vanilla} to denote the correct vanilla gradient estimate; meanwhile, \emph{variance reduced (incorrect)} denotes differentiation through the variance reduced KL estimate. In both cases, we use the label \emph{incorrect} to highlight the fact that these two implementations are technically inaccurate.

\paragraph{Bias of gradient estimates.}
Figure~\ref{figure:kl-gradient} 
 shows the MSE of KL gradient estimate as a function of sample size $n$. As $n$ increases, we see that the MSE decreases for the vanilla estimates at a log scale, while both vanilla (incorrect) and variance reduced gradient estimates asymptote to a higher error, indicating the bias.

\paragraph{Minimizing KL divergence.}
Then, we initialize $\pi$ to be far from a target and carry out iterative updates using the gradient estimates. Figure~\ref{figure:kl-minimization} right plot shows how the target KL divergence is minimized over time. Using the analytic gradient as the gold standard, we see that vanilla estimates behave as expected. 

The vanilla (incorrect) estimate causes the KL divergence to increase over time, likely due to the fact that the update is zero-mean and essentially randomly drifts the parameter around. The variance reduced gradient estimate, though biased, drives down the KL divergence too at a slower rate. This can be explained by the fact that the gradient estimate minimizes a reversed KL divergence that pulls $\pi$ to be close to the target policy as well.

\begin{figure}[t]
\centering
\includegraphics[width=.45\textwidth]{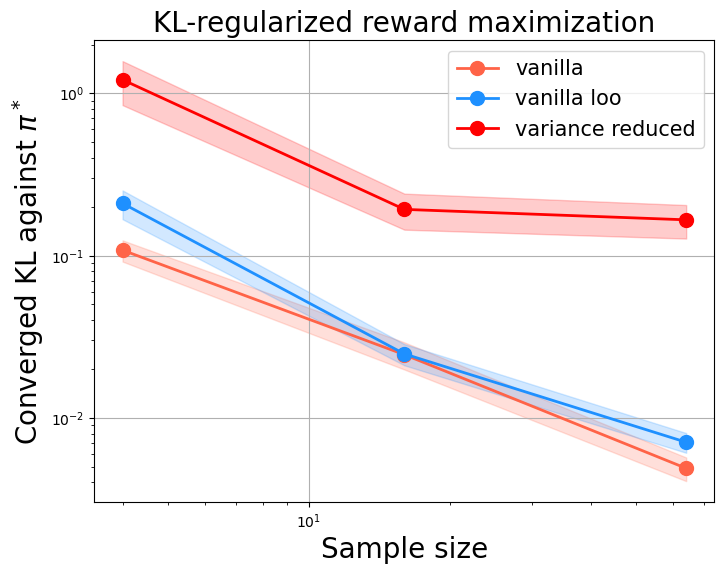}
\caption{\small{The converged value of $\mathbb{KL}(\pi,\pi^*)$ as a function of the sample size. As sample size increases, vanilla gradient estimates obtain lower error values. However, the incorrectly implemented variance reduced estimate plateaus at a higher error.}}
\label{figure:converged-kl}
\end{figure}

\paragraph{KL-regularized reward maximization.} We set up a KL-regularized maximization problem where we initialize $\pi=\pi_\text{ref}$ at the beginning,
\begin{align*}
    \max_\pi \mathbb{E}_{y\sim \pi}\left[r(y)\right] - \beta \mathbb{KL}\left(\pi,\pi_\text{ref}\right).
\end{align*}
The optimal policy is $\pi^*(y)\propto \pi_\text{ref}(y)\exp(\beta^{-1}r(y))$. To maximize the objective, we carry out gradient ascent combining the policy gradient estimate \citep{Sutton98} and a KL gradient estimate. We measure the KL-divergence between $\pi$ and $\pi^*$ over time; in fact, we can see that the above objective itself is a constant away from $-\mathbb{KL}(\pi,\pi^*)$.

Figure~\ref{figure:regularized-kl} 
 shows the normalized KL (normalized by the initial value $\mathbb{KL}(\pi_\text{ref},\pi^*))$ during training. The vanilla gradient estimates improve upon the KL estimates, though converge to a value that depends on the number of samples $n$ used for gradient estimation. Indeed, Figure~\ref{figure:converged-kl} shows that as $n$ increases, vanilla gradient estimates improve in the asymptotic accuracy. Meanwhile, the vanilla (incorrect) gradient estimate and the variance reduced gradient estimate do not improve upon KL against the optimal policy.

\begin{figure}[t]
\centering
\includegraphics[width=.45\textwidth]{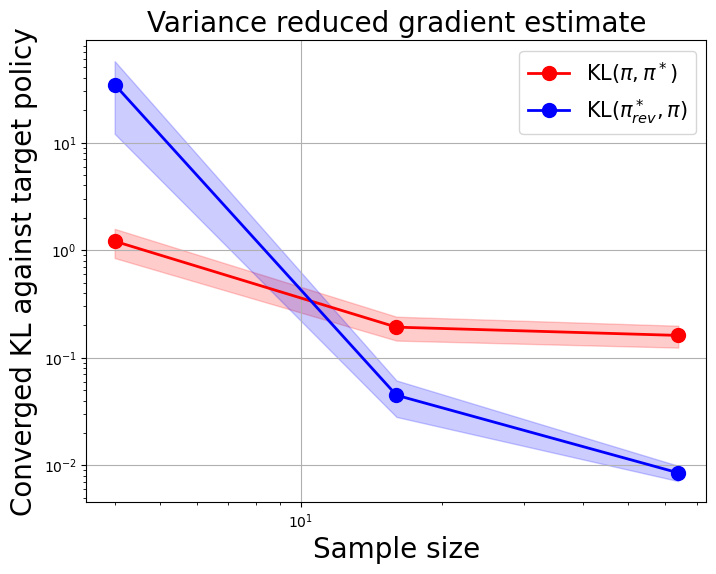}
\caption{\small{When updating with the variance reduced estimate, we measure the divergence between $\pi$ against the supposed target $\pi^*$ and its true target $\pi_\text{rev}^*$. It is clear that variance reduced gradient estimate leads to the convergence to $\pi_\text{rev}^*$ (different from $\pi^*$).}}
\label{figure:converged-metrics}
\end{figure}

\paragraph{Variance reduced gradient estimate converges to a different policy.} As alluded to before, the variance reduced gradient estimate ends up estimating the gradient to the reversed KL divergence. To understand its convergence, we can define another regularized optimization problem
\begin{align*}
    \max_\pi \mathbb{E}_{y\sim \pi}\left[r(y)\right] - \beta \mathbb{KL}\left(\pi_\text{ref},\pi\right).
\end{align*}
using the reversed KL  $\mathbb{KL}\left(\pi_\text{ref},\pi\right) \neq \mathbb{KL}\left(\pi,\pi_\text{ref}\right)$. The optimal policy is $\pi_\text{rev}^*(y)=\frac{\beta\pi_\text{ref}(y)}{\lambda^*-r(y)}$ where $\lambda^*\in\mathbb{R}$ is the dual variable ensuring that the $\pi_\text{rev}^*$ is a probability distribution. We refer to \citet{grill2020monte} for discussion on the property of $\pi_\text{rev}^*$ and how to find $\lambda^*$ with a simple binary search. 

Figure~\ref{figure:converged-metrics}  shows that with increase in the sample size, the converged KL against $\pi_\text{rev}^*$ decreases more smoothly as compared to the plateauing divergence against $\pi^*$, corroborating the fact that here $\pi$ converges to $\pi_\text{rev}^*$ instead of $\pi^*$.

\section{KL divergence and gradient estimate for sequences}

So far our discussions have focused on the KL divergence for a single random variable $y\in \mathcal{Y}$. For sequence modeling applications such as LLM, the variable $y$ denotes a whole sequence of tokens $y_{1:T}$ with random length $T$. The \emph{sequence-level} KL divergence is the expected sum of \emph{token-level} KLs, 
\begin{align*} 
    \mathbb{KL}_\text{seq}(\pi,\pi_\text{ref}) = \mathbb{E}_{\pi}\left[\sum_{t=1}^T \mathbb{KL}\left(\pi\left(\cdot\;|\; y_{1:t-1}\right),\pi_\text{ref}\left(\cdot\;|\; y_{1:t-1}\right)\right)\right], 
\end{align*}
where $\pi
    \left(\cdot \;|\; y_{1:t-1}\right)$ denotes the auto-regressive conditional distribution over next token $y_t$. The expectation is defined for $y$ being  sampled auto-regressively from $\pi$.

A popular way to implement minimization sequence-level KL divergence, is to define token-level KL losses and compute descend on its gradient \citep{vonwerra2022trl,lambert2024tulu3}. Concretely, the aggregate gradient estimate is
\begin{align} 
    \sum_{t=1}^T \nabla \widehat{\mathbb{KL}}\left(\pi\left(\cdot\;|\; y_{1:t-1}\right),\pi_\text{ref}\left(\cdot\;|\; y_{1:t-1}\right)\right),\label{eq:partial}
\end{align}
where $\nabla \widehat{\mathbb{KL}}\left(\pi\left(\cdot\;|\; y_{1:t-1}\right),\pi_\text{ref}\left(\cdot\;|\; y_{1:t-1}\right)\right)$ can be computed using various estimates from previous discussions, using the sampled token $y_t\sim \pi\left(\cdot\;|\; y_{1:t-1}\right)$.

\subsection{Token-level loss does not produce proper gradient}

Noticeably, even when the token-level loss implements the unbiased gradient estimate to the token-level KL (e.g., using Eqn~\eqref{eq:vanilla-correct}), the above sequence-level estimate is not an unbiased estimate to the sequence-level KL gradient. We refer readers to \citet{tang2025rl} for extended discussions. Importantly, the above gradient does not account for the impact that token $y_t$ has on the future tokens $y_s,s>t$ which impacts future token-level KL divergences. Henceforth, we call the above loss implementation the \emph{token-level} losses, which constitute estimates to partial gradient of sequence-level KL divergence, see Appendix~\ref{appendix:sequence-kl} for more elaborations.

We now examine the special case where the token-level loss is implemented as the variance reduced gradient estimate (Eqn~\eqref{eq:vanilla-correct}), where the expectation of the gradient estimate is
\begin{align*}
    \mathbb{E}_{y\sim \pi}\left[\sum_{t=1}^T \nabla \mathbb{KL}\left(\pi_\text{ref}\left(\cdot\;|\; y_{1:t-1}\right),\pi\left(\cdot\;|\; y_{1:t-1}\right)\right)\right].
\end{align*}
Though this bears some similarity to the gradient of the sequence-level reverse KL $\mathbb{KL}_\text{seq}(\pi_\text{ref},\pi)$, the key difference is that the latter is defined through sequences $y$ sampled from $\pi_\text{ref}$ rather than $\pi$. As a result, a variance reduced estimate based token-level loss does not produce even the partial gradient of any existing sequence-level divergence.

\mdfsetup{%
backgroundcolor=black!10,
roundcorner=10pt}
\begin{mdframed}
\textbf{Key insight.}\\
Token-level losses only produce a partial derivative at best to the full sequence-level KL divergence.
\end{mdframed}

\subsection{Unbiased estimate to the full sequence gradient}
There are a few ways to estimate the full sequence gradient that accounts for the impact token $y_t$ has on future steps $s>t$. Let $\rho_t$ denote the ratio $\pi(y_t\;|\;y_{1:t-1}) / \pi_\text{ref}(y_t\;|\;y_{1:t-1})$. We start with the sequence-level vanilla estimate 
\begin{align}
\left(\sum_{t=1}^T \log \rho_t \right)\left(\sum_{t=1}^T \nabla \log \pi\left(y_t\;|\; y_{1:t-1}\right)\right)\label{eq:seq-vanilla}
\end{align}
which can be derived from applying the vanilla gradient estimate formula to the sequence as a single entity.
Starting from the vanilla estimate, there are two complementary ways to reduce variance. One way is to make use of leave-one-out control variate at the sequence level, given there are $n>1$ sequences in a batch, 
\begin{align*}
\frac{1}{n}\sum_{i=1}^n\left(\sum_{t=1}^{T_i} \log \rho_{i,t} - \tilde{v}_i \right)\left(\sum_{t=1}^{T_i} \nabla \log \pi\left(y_{i,t}\;|\; y_{i,1:t-1}\right)\right),
\end{align*}
where $\tilde{v}_i=\frac{1}{n-1}\sum_{j\neq i}\sum_{t=1}^{T_j} \log \rho_{j,t}$ is the control variate. 

Alternatively, the variance of the estimate can be further reduced by noticing that $\E\left[\log  \rho_s \nabla \log\pi(y_t \;|\; y_{1:t-1})\right] = \E_{y_{1:t-1}}\left[ \log \rho_s \E_{y_t\sim \pi(\cdot \;|\; y_{1:t-1})}\left[\nabla \log\pi(y_t \;|\; y_{1:t-1})\right]\right]=0$ for time step $s<t$. Thus for better credit assignment, we can remove almost half of the terms in the previous estimate:
\begin{align*}
    \sum_{t=1}^T \left(\sum_{s=t}^T\log \rho_s\right)\nabla \log \pi\left(y_t\;|\; y_{1:t-1}\right).
\end{align*}
We call this variant \emph{cumulative} estimate, indicating the fact that $\sum_{s=t}^T\log \rho_s$ is reminiscent of cumulative reward in RL. However, because it is difficult to compute useful control variate at the token level, it is hard to improve upon the cumulative estimate with other variance reduction methods.

\paragraph{Analytic gradient.} Note that computing the analytic gradient (Eqn~\eqref{eq:analytic}) to the sequence-level KL is not feasible, because the sample space $\mathcal{Y}$ corresponds to all possible sequences of size $V^{T_\text{max}}$ where $V$ is the per-token vocab size and $T_\text{max}$ the maximum sequence length. Though it is possible to compute per-token losses analytically, where the sample space is of size $V$, this does not account for the temporal dependencies of different time steps discussed above.

\section{LLM experiments}

We investigate the impact that various token-loss implementations and sequence-level gradient estimates have on the practical performance of LLM applications.

\subsection{Low-KL regime: KL-regularized reward maximization}

We initialize $\pi=\pi_\text{ref}$ as the 8B Llama 3 model \citep{grattafiori2024llama} and train on MATH training prompts with string match reward \citep{hendrycks2021measuring}. We measure the training performance as a function of the sequence-level KL divergence $\mathbb{KL}(\pi,\pi_\text{ref})$. With stronger regularization (larger value of $\beta$), we observe that the training becomes more KL-efficient \citep{gao2023scaling}, i.e., the same training reward can be obtained at lower KL.

\begin{figure}[t]
\centering
\includegraphics[width=.45\textwidth]{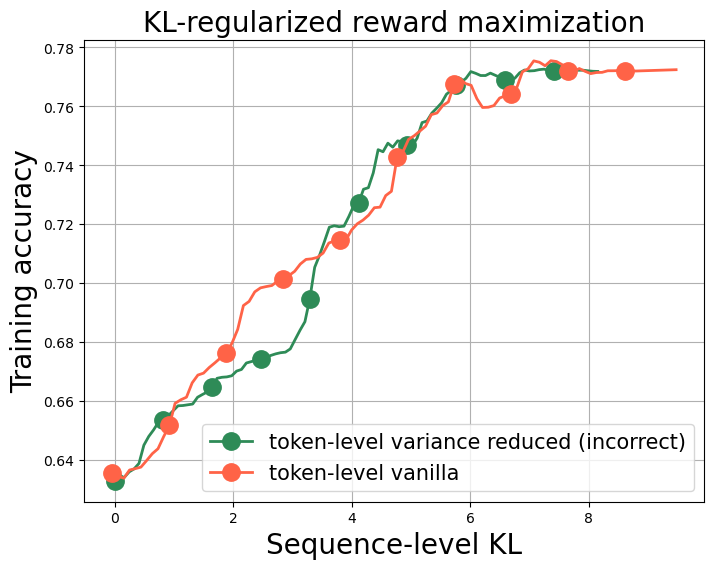}
\caption{\small{KL-regularized reward maximization with 8B models. Different estimates behave similarly in the low-KL regime, where enforcing different divergences lead to similar performance-KL trade-off.}}
\label{figure:kl-reward}
\end{figure}

Since the vanilla (incorrect) estimate does not enforce KL regularization at all, it is the least KL efficient regardless of the value of $\beta$. Figure~\ref{figure:kl-reward} compares the performance of variance reduced (incorrect) estimate and the vanilla estimate, both at $\beta=0.001$. In general, we find the two estimates to trace out a similar KL-performance curve at different $\beta$. We speculate that this is because the training is at low-KL regime where various divergences have similar behavior. In other words, $\mathbb{KL}(\pi,\pi_\text{ref})\approx \mathbb{KL}(\pi_\text{ref},\pi)$, and enforcing  KL vs. the reverse KL does not make a big difference.

However, we speculate that for long-run RL training  \citep{jaech2024openai,guo2025deepseek} where $\pi$ is allowed to deviate significantly from $\pi_\text{ref}$, different divergences might make a difference in the limit. We leave this for further experiments.

\mdfsetup{%
backgroundcolor=black!10,
roundcorner=10pt}
\begin{mdframed}
\textbf{Key insight.}\\
Different divergences do not make a significant practical difference in the low-KL regime, as is often the case with strongly regularized reward maximization.
\end{mdframed}

\begin{figure}[t]
\centering
\includegraphics[width=.45\textwidth]{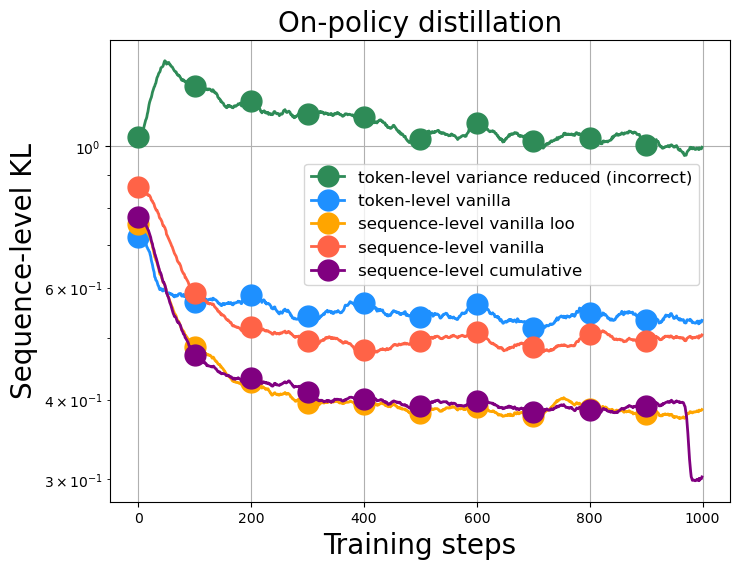}
\caption{\small{On-policy distillation of a 70B policy $\pi_\text{ref}$ into a 8B policy $\pi$ by minimizing KL divergence. To compare token-level losses, the vanilla gradient estimate  minimizes the target metric, while the variance reduced estimate does not. Furthermore, proper sequence-level implementations minimize the KL at faster rate, with significant impact from variance reduction techniques.}}
\label{figure:kl-distill}
\end{figure}

\subsection{High-KL regime: on-policy distillation}

Next, we study a case where the goal is to distill information from $\pi_\text{ref}$ to $\pi$ through on-policy distillation \citep{agarwal2024onpolicydistillationlanguagemodels}, by minimizing the KL divergence. For this, we initialize $\pi_\text{ref}$ to be a 70B model and $\pi$ a 8B model. With the distillation case, we emulate applications in the high-KL regime, and study the behavior of different algorithms.

Figure~\ref{figure:kl-distill} shows the sequence-level KL divergence over time. First, we compare the token-level losses: while the vanilla estimate  decreases the KL divergence over time, the variance reduced (incorrect) estimate does not make much progress on the target metric. This is meant to illustrate that the two estimates behave differently when $\pi$ is far from $\pi_\text{ref}$.

\subsection{Impact of unbiased sequence-level gradient estimate}

To see the impact of the the sequence-level KL gradient, we further examine Figure~\ref{figure:kl-distill}. Note that when the gradient is implemented in an unbiased way, the vanilla sequence-level gradient estimate (Eqn~\eqref{eq:seq-vanilla}) improves over the best token-level implementations. Furthermore, the two variants that apply variance reduction (leave-one-out and cumulative estimates) both further improve the rate of learning.

Taken together, we see that a properly implemented sequence-level KL gradient improves upon the efficacy of minimizing the sequence-level KL, especially in high-KL regime.

\mdfsetup{%
backgroundcolor=black!10,
roundcorner=10pt}
\begin{mdframed}
\textbf{Key insight.}\\
Both correctly implemented token-level KL gradient estimates and sequence-level KL gradient estimates impact performance significantly in high-KL regime.
\end{mdframed}

\section{Discussion and related work}

We briefly discuss a few aspects of related work.

\paragraph{Stochastic gradient estimation.}
Estimating gradient with samples is well-studied in multiple domains, including financial engineering \citep{glasserman2004monte}, probabilistic inference \citep{kingma2013auto,rezende2014stochastic,ranganath2014black} and RL \citep{williams1992simple,sutton1999policy}. As with KL, such applications apply scalar objectives that depend on learnable parameters through a sampling process, hence the gradient estimation is more sophisticated and distinct from supervised learning \citep{schulman2015gradient}.

\paragraph{Design space of regularizations.}
Though the canonical formulation of regularization policy optimization adopts the KL divergence $\mathbb{KL}(\pi,\pi_\text{ref})$, recent work has explored alternative divergences \citep{azar2024general,tang2024generalized, huang2024correcting}. Concurrently, \citet{zhang2025design} discussed the designs of regularizations and noted that the variance reduced estimate enforced the reverse KL divergence, similar to our first pitfall. Our messages differ: we identify such implementations as a  \emph{bug} rather than a design.

\paragraph{Off-policy learning.} KL estimation requires sampling from $\pi$. However, in generic off-policy learning where the samples are from a behavior policy $\mu$ different from $\pi$, estimating KL divergence is challenging. In such cases, the squared regularization $\mathbb{E}_{y\sim \mu}\left[\left(\log \frac{\pi(y)}{\pi_\text{ref}(y)}\right)^2\right]$ is the unique regularization that keeps the optimal policy intact \citep{azar2024general,flet2024contrastive,cohen2025soft} and can be implemented faithfully with the sequence formulation \citep{tang2025rl}. Intriguingly, the biased KL estimate in Eqn~\eqref{eq:biased} can be understood as a special case of such regularization in the on-policy case $\mu=\pi$, hinting at that it is the unique loss which produces the correct gradient.

\section{Conclusion}

We have studied a few implementations of KL gradient in the form of loss functions, which do not produce the targeted gradient in expectation. Interestingly, we find that the variance reduced estimate as referenced in a few recent work, ends up producing a reverse KL regularization. Yet,  it is important to note that the resulting regularization is rather incidental, 
and does not produce desirable behavior in general especially in the high-KL regime.

\bibliography{your_bib_file}

\begin{thebibliography}{41}
\providecommand{\natexlab}[1]{#1}
\providecommand{\url}[1]{\texttt{#1}}
\expandafter\ifx\csname urlstyle\endcsname\relax
  \providecommand{\doi}[1]{doi: #1}\else
  \providecommand{\doi}{doi: \begingroup \urlstyle{rm}\Url}\fi

\bibitem[Agarwal et~al.(2024)Agarwal, Vieillard, Zhou, Stanczyk, Ramos, Geist,
  and Bachem]{agarwal2024onpolicydistillationlanguagemodels}
Rishabh Agarwal, Nino Vieillard, Yongchao Zhou, Piotr Stanczyk, Sabela Ramos,
  Matthieu Geist, and Olivier Bachem.
\newblock On-policy distillation of language models: Learning from
  self-generated mistakes, 2024.
\newblock URL \url{https://arxiv.org/abs/2306.13649}.

\bibitem[Azar et~al.(2024)Azar, Guo, Piot, Munos, Rowland, Valko, and
  Calandriello]{azar2024general}
Mohammad~Gheshlaghi Azar, Zhaohan~Daniel Guo, Bilal Piot, Remi Munos, Mark
  Rowland, Michal Valko, and Daniele Calandriello.
\newblock A general theoretical paradigm to understand learning from human
  preferences.
\newblock In \emph{International Conference on Artificial Intelligence and
  Statistics}, pages 4447--4455. PMLR, 2024.

\bibitem[Bradbury et~al.(2018)Bradbury, Frostig, Hawkins, Johnson, Leary,
  Maclaurin, Necula, Paszke, Vander{P}las, Wanderman-{M}ilne, and
  Zhang]{jax2018github}
James Bradbury, Roy Frostig, Peter Hawkins, Matthew~James Johnson, Chris Leary,
  Dougal Maclaurin, George Necula, Adam Paszke, Jake Vander{P}las, Skye
  Wanderman-{M}ilne, and Qiao Zhang.
\newblock {JAX}: composable transformations of {P}ython+{N}um{P}y programs,
  2018.
\newblock URL \url{http://github.com/jax-ml/jax}.

\bibitem[Christiano et~al.(2017)Christiano, Leike, Brown, Martic, Legg, and
  Amodei]{christiano2017deep}
Paul~F Christiano, Jan Leike, Tom Brown, Miljan Martic, Shane Legg, and Dario
  Amodei.
\newblock Deep reinforcement learning from human preferences.
\newblock \emph{Advances in neural information processing systems}, 30, 2017.

\bibitem[Cohen et~al.(2025)Cohen, Zhang, Zheng, Tang, Munos, and
  Synnaeve]{cohen2025soft}
Taco Cohen, David~W Zhang, Kunhao Zheng, Yunhao Tang, Remi Munos, and Gabriel
  Synnaeve.
\newblock Soft policy optimization: Online off-policy rl for sequence models.
\newblock \emph{arXiv preprint arXiv:2503.05453}, 2025.

\bibitem[Flet-Berliac et~al.(2024)Flet-Berliac, Grinsztajn, Strub, Wu, Choi,
  Cremer, Ahmadian, Chandak, Azar, Pietquin, et~al.]{flet2024contrastive}
Yannis Flet-Berliac, Nathan Grinsztajn, Florian Strub, Bill Wu, Eugene Choi,
  Chris Cremer, Arash Ahmadian, Yash Chandak, Mohammad~Gheshlaghi Azar, Olivier
  Pietquin, et~al.
\newblock Contrastive policy gradient: Aligning llms on sequence-level scores
  in a supervised-friendly fashion.
\newblock \emph{arXiv preprint arXiv:2406.19185}, 2024.

\bibitem[Gao et~al.(2023)Gao, Schulman, and Hilton]{gao2023scaling}
Leo Gao, John Schulman, and Jacob Hilton.
\newblock Scaling laws for reward model overoptimization.
\newblock In \emph{International Conference on Machine Learning}, pages
  10835--10866. PMLR, 2023.

\bibitem[Glasserman(2004)]{glasserman2004monte}
P~Glasserman.
\newblock Monte carlo methods in financial engineering, 2004.

\bibitem[Grattafiori et~al.(2024)Grattafiori, Dubey, Jauhri, Pandey, Kadian,
  Al-Dahle, Letman, Mathur, Schelten, Vaughan, et~al.]{grattafiori2024llama}
Aaron Grattafiori, Abhimanyu Dubey, Abhinav Jauhri, Abhinav Pandey, Abhishek
  Kadian, Ahmad Al-Dahle, Aiesha Letman, Akhil Mathur, Alan Schelten, Alex
  Vaughan, et~al.
\newblock The llama 3 herd of models.
\newblock \emph{arXiv preprint arXiv:2407.21783}, 2024.

\bibitem[Grill et~al.(2020)Grill, Altch{\'e}, Tang, Hubert, Valko, Antonoglou,
  and Munos]{grill2020monte}
Jean-Bastien Grill, Florent Altch{\'e}, Yunhao Tang, Thomas Hubert, Michal
  Valko, Ioannis Antonoglou, and R{\'e}mi Munos.
\newblock Monte-carlo tree search as regularized policy optimization.
\newblock In \emph{International Conference on Machine Learning}, pages
  3769--3778. PMLR, 2020.

\bibitem[Guo et~al.(2025)Guo, Yang, Zhang, Song, Zhang, Xu, Zhu, Ma, Wang, Bi,
  et~al.]{guo2025deepseek}
Daya Guo, Dejian Yang, Haowei Zhang, Junxiao Song, Ruoyu Zhang, Runxin Xu,
  Qihao Zhu, Shirong Ma, Peiyi Wang, Xiao Bi, et~al.
\newblock Deepseek-r1: Incentivizing reasoning capability in llms via
  reinforcement learning.
\newblock \emph{arXiv preprint arXiv:2501.12948}, 2025.

\bibitem[Haarnoja et~al.(2018)Haarnoja, Zhou, Abbeel, and
  Levine]{haarnoja2018soft}
Tuomas Haarnoja, Aurick Zhou, Pieter Abbeel, and Sergey Levine.
\newblock Soft actor-critic: Off-policy maximum entropy deep reinforcement
  learning with a stochastic actor.
\newblock In \emph{International conference on machine learning}, pages
  1861--1870. PMLR, 2018.

\bibitem[Hendrycks et~al.(2021)Hendrycks, Burns, Kadavath, Arora, Basart, Tang,
  Song, and Steinhardt]{hendrycks2021measuring}
Dan Hendrycks, Collin Burns, Saurav Kadavath, Akul Arora, Steven Basart, Eric
  Tang, Dawn Song, and Jacob Steinhardt.
\newblock Measuring mathematical problem solving with the math dataset.
\newblock \emph{arXiv preprint arXiv:2103.03874}, 2021.

\bibitem[Huang et~al.(2024)Huang, Zhan, Xie, Lee, Sun, Krishnamurthy, and
  Foster]{huang2024correcting}
Audrey Huang, Wenhao Zhan, Tengyang Xie, Jason~D Lee, Wen Sun, Akshay
  Krishnamurthy, and Dylan~J Foster.
\newblock Correcting the mythos of kl-regularization: Direct alignment without
  overoptimization via chi-squared preference optimization.
\newblock \emph{arXiv preprint arXiv:2407.13399}, 2024.

\bibitem[Jaech et~al.(2024)Jaech, Kalai, Lerer, Richardson, El-Kishky, Low,
  Helyar, Madry, Beutel, Carney, et~al.]{jaech2024openai}
Aaron Jaech, Adam Kalai, Adam Lerer, Adam Richardson, Ahmed El-Kishky, Aiden
  Low, Alec Helyar, Aleksander Madry, Alex Beutel, Alex Carney, et~al.
\newblock Openai o1 system card.
\newblock \emph{arXiv preprint arXiv:2412.16720}, 2024.

\bibitem[Kingma et~al.(2013)Kingma, Welling, et~al.]{kingma2013auto}
Diederik~P Kingma, Max Welling, et~al.
\newblock Auto-encoding variational bayes, 2013.

\bibitem[Kool et~al.(2019)Kool, van Hoof, and Welling]{kool2019buy}
Wouter Kool, Herke van Hoof, and Max Welling.
\newblock Buy 4 reinforce samples, get a baseline for free!
\newblock 2019.

\bibitem[Lambert et~al.(2024)Lambert, Morrison, Pyatkin, Huang, Ivison,
  Brahman, Miranda, Liu, Dziri, Lyu, Gu, Malik, Graf, Hwang, Yang, Bras,
  Tafjord, Wilhelm, Soldaini, Smith, Wang, Dasigi, and
  Hajishirzi]{lambert2024tulu3}
Nathan Lambert, Jacob Morrison, Valentina Pyatkin, Shengyi Huang, Hamish
  Ivison, Faeze Brahman, Lester James~V. Miranda, Alisa Liu, Nouha Dziri, Shane
  Lyu, Yuling Gu, Saumya Malik, Victoria Graf, Jena~D. Hwang, Jiangjiang Yang,
  Ronan~Le Bras, Oyvind Tafjord, Chris Wilhelm, Luca Soldaini, Noah~A. Smith,
  Yizhong Wang, Pradeep Dasigi, and Hannaneh Hajishirzi.
\newblock Tülu 3: Pushing frontiers in open language model post-training.
\newblock 2024.

\bibitem[Liu et~al.(2025)Liu, Chen, Li, Qi, Pang, Du, Lee, and
  Lin]{liu2025understanding}
Zichen Liu, Changyu Chen, Wenjun Li, Penghui Qi, Tianyu Pang, Chao Du, Wee~Sun
  Lee, and Min Lin.
\newblock Understanding r1-zero-like training: A critical perspective.
\newblock \emph{arXiv preprint arXiv:2503.20783}, 2025.

\bibitem[Mnih and Rezende(2016)]{mnih2016variational}
Andriy Mnih and Danilo Rezende.
\newblock Variational inference for monte carlo objectives.
\newblock In \emph{International Conference on Machine Learning}, pages
  2188--2196. PMLR, 2016.

\bibitem[Ouyang et~al.(2022)Ouyang, Wu, Jiang, Almeida, Wainwright, Mishkin,
  Zhang, Agarwal, Slama, Ray, et~al.]{ouyang2022training}
Long Ouyang, Jeffrey Wu, Xu~Jiang, Diogo Almeida, Carroll Wainwright, Pamela
  Mishkin, Chong Zhang, Sandhini Agarwal, Katarina Slama, Alex Ray, et~al.
\newblock Training language models to follow instructions with human feedback.
\newblock \emph{Advances in Neural Information Processing Systems},
  35:\penalty0 27730--27744, 2022.

\bibitem[Paszke et~al.(1912)Paszke, Gross, Massa, Lerer, Bradbury, Chanan,
  Killeen, Lin, Gimelshein, Antiga, et~al.]{paszke1912pytorch}
Adam Paszke, Sam Gross, Francisco Massa, Adam Lerer, James Bradbury, Gregory
  Chanan, Trevor Killeen, Zeming Lin, Natalia Gimelshein, Luca Antiga, et~al.
\newblock Pytorch: An imperative style, high-performance deep learning library.
  arxiv 2019.
\newblock \emph{arXiv preprint arXiv:1912.01703}, 10, 1912.

\bibitem[Ranganath et~al.(2014)Ranganath, Gerrish, and
  Blei]{ranganath2014black}
Rajesh Ranganath, Sean Gerrish, and David Blei.
\newblock Black box variational inference.
\newblock In \emph{Artificial intelligence and statistics}, pages 814--822.
  PMLR, 2014.

\bibitem[Rezende et~al.(2014)Rezende, Mohamed, and
  Wierstra]{rezende2014stochastic}
Danilo~Jimenez Rezende, Shakir Mohamed, and Daan Wierstra.
\newblock Stochastic backpropagation and approximate inference in deep
  generative models.
\newblock In \emph{International conference on machine learning}, pages
  1278--1286. PMLR, 2014.

\bibitem[Robert et~al.(1999)Robert, Casella, and Casella]{robert1999monte}
Christian~P Robert, George Casella, and George Casella.
\newblock \emph{Monte Carlo statistical methods}, volume~2.
\newblock Springer, 1999.

\bibitem[Schulman(2017)]{schulman-blogpost}
John Schulman.
\newblock kl-approx blogpost.
\newblock \url{https://joschu.net/blog/kl-approx.html}, 2017.

\bibitem[Schulman et~al.(2015{\natexlab{a}})Schulman, Heess, Weber, and
  Abbeel]{schulman2015gradient}
John Schulman, Nicolas Heess, Theophane Weber, and Pieter Abbeel.
\newblock Gradient estimation using stochastic computation graphs.
\newblock \emph{Advances in neural information processing systems}, 28,
  2015{\natexlab{a}}.

\bibitem[Schulman et~al.(2015{\natexlab{b}})Schulman, Levine, Abbeel, Jordan,
  and Moritz]{schulman2015trust}
John Schulman, Sergey Levine, Pieter Abbeel, Michael Jordan, and Philipp
  Moritz.
\newblock Trust region policy optimization.
\newblock In \emph{International conference on machine learning}, pages
  1889--1897. PMLR, 2015{\natexlab{b}}.

\bibitem[Schulman et~al.(2017{\natexlab{a}})Schulman, Chen, and
  Abbeel]{schulman2017equivalence}
John Schulman, Xi~Chen, and Pieter Abbeel.
\newblock Equivalence between policy gradients and soft q-learning.
\newblock \emph{arXiv preprint arXiv:1704.06440}, 2017{\natexlab{a}}.

\bibitem[Schulman et~al.(2017{\natexlab{b}})Schulman, Wolski, Dhariwal,
  Radford, and Klimov]{schulman2017proximal}
John Schulman, Filip Wolski, Prafulla Dhariwal, Alec Radford, and Oleg Klimov.
\newblock Proximal policy optimization algorithms.
\newblock \emph{arXiv preprint arXiv:1707.06347}, 2017{\natexlab{b}}.

\bibitem[Shao et~al.(2024)Shao, Wang, Zhu, Xu, Song, Bi, Zhang, Zhang, Li, Wu,
  et~al.]{shao2024deepseekmath}
Zhihong Shao, Peiyi Wang, Qihao Zhu, Runxin Xu, Junxiao Song, Xiao Bi, Haowei
  Zhang, Mingchuan Zhang, YK~Li, Y~Wu, et~al.
\newblock Deepseekmath: Pushing the limits of mathematical reasoning in open
  language models.
\newblock \emph{arXiv preprint arXiv:2402.03300}, 2024.

\bibitem[Sutton and Barto(1998)]{Sutton98}
R.~Sutton and A.~Barto.
\newblock \emph{Reinforcement Learning: An Introduction}.
\newblock {MIT} Press, 1998.

\bibitem[Sutton et~al.(1999)Sutton, McAllester, Singh, and
  Mansour]{sutton1999policy}
Richard~S Sutton, David McAllester, Satinder Singh, and Yishay Mansour.
\newblock Policy gradient methods for reinforcement learning with function
  approximation.
\newblock \emph{Advances in neural information processing systems}, 12, 1999.

\bibitem[Tang et~al.(2024)Tang, Guo, Zheng, Calandriello, Munos, Rowland,
  Richemond, Valko, Pires, and Piot]{tang2024generalized}
Yunhao Tang, Zhaohan~Daniel Guo, Zeyu Zheng, Daniele Calandriello, R{\'e}mi
  Munos, Mark Rowland, Pierre~Harvey Richemond, Michal Valko,
  Bernardo~{\'A}vila Pires, and Bilal Piot.
\newblock Generalized preference optimization: A unified approach to offline
  alignment.
\newblock \emph{arXiv preprint arXiv:2402.05749}, 2024.

\bibitem[Tang et~al.(2025)Tang, Cohen, Zhang, Valko, and Munos]{tang2025rl}
Yunhao Tang, Taco Cohen, David~W Zhang, Michal Valko, and R{\'e}mi Munos.
\newblock Rl-finetuning llms from on-and off-policy data with a single
  algorithm.
\newblock \emph{arXiv preprint arXiv:2503.19612}, 2025.

\bibitem[von Werra et~al.(2020)von Werra, Belkada, Tunstall, Beeching, Thrush,
  Lambert, Huang, Rasul, and Gallouédec]{vonwerra2022trl}
Leandro von Werra, Younes Belkada, Lewis Tunstall, Edward Beeching, Tristan
  Thrush, Nathan Lambert, Shengyi Huang, Kashif Rasul, and Quentin Gallouédec.
\newblock Trl: Transformer reinforcement learning.
\newblock \url{https://github.com/huggingface/trl}, 2020.

\bibitem[Williams(1992)]{williams1992simple}
Ronald~J Williams.
\newblock Simple statistical gradient-following algorithms for connectionist
  reinforcement learning.
\newblock \emph{Machine learning}, 8:\penalty0 229--256, 1992.

\bibitem[Zhang et~al.(2025)Zhang, Liu, Yuan, Yuan, Gu, and
  Yao]{zhang2025design}
Yifan Zhang, Yifeng Liu, Huizhuo Yuan, Yang Yuan, Quanquan Gu, and Andrew~C
  Yao.
\newblock On the design of kl-regularized policy gradient algorithms for llm
  reasoning.
\newblock \emph{arXiv preprint arXiv:2505.17508}, 2025.

\bibitem[Zhao et~al.(2025)Zhao, Gupta, Zheng, and Grover]{zhao2025d1}
Siyan Zhao, Devaansh Gupta, Qinqing Zheng, and Aditya Grover.
\newblock d1: Scaling reasoning in diffusion large language models via
  reinforcement learning.
\newblock \emph{arXiv preprint arXiv:2504.12216}, 2025.

\bibitem[Ziebart et~al.(2008)Ziebart, Maas, Bagnell, Dey,
  et~al.]{ziebart2008maximum}
Brian~D Ziebart, Andrew~L Maas, J~Andrew Bagnell, Anind~K Dey, et~al.
\newblock Maximum entropy inverse reinforcement learning.
\newblock In \emph{Aaai}, volume~8, pages 1433--1438. Chicago, IL, USA, 2008.

\bibitem[Ziegler et~al.(2019)Ziegler, Stiennon, Wu, Brown, Radford, Amodei,
  Christiano, and Irving]{ziegler2019fine}
Daniel~M Ziegler, Nisan Stiennon, Jeffrey Wu, Tom~B Brown, Alec Radford, Dario
  Amodei, Paul Christiano, and Geoffrey Irving.
\newblock Fine-tuning language models from human preferences.
\newblock \emph{arXiv preprint arXiv:1909.08593}, 2019.

\end{thebibliography}
\bibliographystyle{plainnat}

\appendix
\newpage
\section{Details on theoretical results of variance reduced gradient estimate}
\label{appendix:derivation}
We provide some more details on the theoretical results of the variance reduced gradient estimate. It follows that 
\begin{align*}
    \mathbb{E}_{y\sim \pi}\left[\widehat{\nabla \mathbb{KL}}_\text{var-reduced}\right] &=_{(a)}\mathbb{E}_{y\sim \pi}\left[-\frac{\pi_\text{ref}(y)}{\pi(y)^2}\nabla \pi(y)\right] \\
    &=_{(b)} \mathbb{E}_{y\sim \pi}\left[-\frac{\pi_\text{ref}(y)}{\pi(y)}\nabla \log \pi(y)\right]  \\
    &=_{(c)} -\mathbb{E}_{y\sim \pi_\text{ref}}\left[-\nabla \log \pi(y)\right] \\
    &=_{(d)} \nabla \mathbb{KL}(\pi_\text{ref},\pi).
\end{align*}
Here, (a) is due to the fact that the other two terms in the gradient estimate produce zero expectation; (b) is due to the fact that $\nabla \log \pi(y) = \frac{\nabla \pi(y)}{\pi(y)}$; (c) makes use of importance sampling. Finally, (d) comes from the definition of the reverse KL divergence. 

\section{Discussion on the two parts of KL gradient}
\label{appendix:gradient}

Given an unbiased KL estimate $\widehat{\mathbb{KL}}$, we argue that the correct gradient estimate is
\begin{align*}
    \widehat{g} = \underbrace{\nabla \widehat{\mathbb{KL}}}_{\text{path-wise derivative}} + \underbrace{\widehat{\mathbb{KL}} \cdot \nabla \log \pi(y)}_{\text{score function derivative}}.
\end{align*}
Indeed, we see that since $\mathbb{KL}$ depends on $\pi$ both in terms of the sampling distribution and the integrand, its gradient is written as
\begin{align*}
    \nabla \mathbb{KL} &= \nabla \mathbb{E}_{y\sim \pi}\left[\widehat{\mathbb{KL}}\right] \\
    &= \mathbb{E}_{y\sim \pi}\left[\nabla \widehat{\mathbb{KL}}\right] + \mathbb{E}_{y\sim \pi}\left[\widehat{\mathbb{KL}} \cdot \nabla \log \pi(y)\right],
\end{align*}
corresponding to the two parts of the gradient estimates alluded to earlier \citep{glasserman2004monte}. It is easy to see that  $\mathbb{E}\left[\hat{g}\right] =  \nabla \mathbb{KL}$ is indeed the unbiased gradient estimate.

In general, the score function derivative part is non-trivial and can even be just the full gradient (e.g., for the vanilla KL estimate). As a result, the implementation of using $\nabla \widehat{\mathbb{KL}}$ as the gradient estimate, which amounts to just the path-wise derivative part of the gradient, produces a highly biased gradient estimate in all the cases we considered. 

\section{Details on the tabular experiments}\label{appendix:tabular}

Throughout, we carry out tabular experiments with multi-arm bandits. We fix the number of arms $|\mathcal{Y}|=100$ and repeat all experiments with $100$ times to compute the mean and standard errors across independent runs.

For all experiments, we initialize the logits for $\pi_\text{ref}$ and $\pi$ as
\begin{align*}
    \text{logits}(\pi_\text{ref}) = \epsilon_1, \text{logits}(\pi) = \epsilon_1 + \epsilon_2,
\end{align*}
where $\epsilon_i$s are isotropic unit Gaussian vectors. For the MSE measurement experiments, we compute the KL divergence MSE as
\begin{align*}
    \text{MSE} = \left(\widehat{\mathbb{KL}}(\pi,\pi_\text{ref}) - \mathbb{KL}(\pi,\pi_\text{ref}) \right)^2.
\end{align*}
The bias and variance fo the estimates are defined in a conventional way with a fixed sample size $n$. For the gradient measurement experiments, we compute the root MSE as average over parameter dimension $i$,
\begin{align*}
    \text{root-MSE} = \sqrt{\sum_i \left(\widehat{g}_i - \nabla_i \mathbb{KL}(\pi,\pi_\text{ref})\right)^2},
\end{align*}
where $\widehat{g}$ is any particular gradient estimate.

For all optimization problems, $\pi$ is parameterized via softmax parameterization $\pi_t=\text{softmax}(\theta_t)$ per iteration $t$, and $\theta_t$ is updated as
\begin{align*}
\theta_{t+1} = \theta_t - \widehat{g}_t.
\end{align*}
By default, we use $n=4$ samples to estimate gradient $\widehat{g}_t$ except for ablations where $n\in\{16,64\}$. 

A similar implementation holds for KL-regularized reward maximization. We initialize the reward as $|\mathcal{Y}|-$dimensional isotropic Gaussian vectors. The update proceeds as
\begin{align*}
\theta_{t+1} = \theta_t + \widehat{l}_t - \beta \widehat{g}_t.
\end{align*}
Here $\widehat{l}_t=\frac{1}{n}\sum_{i=1}^n (r_i - w_i) \nabla \log \pi_t(y_i|x)$ where $r_i$ is the sampled reward for generation $y_i$. The control variate $w_i = \frac{1}{n-1}\sum_{j\neq i} r_j$ is also based on leave-one-out. Note the gradient estimate for the reward part $\widehat{l}$ is identical across different KL baselines. The analytic gradient baseline applies the exact gradient, which is available upon access to the ground truth reward vector $r$ and $\pi_\text{ref}$. Throughput we adopt a learning rate of $\eta=1$.

\section{Additional discussion on sequence-level KL gradient}\label{appendix:sequence-kl}

 For simplicity of notation, we let $\rho_t\coloneqq \log\frac{\pi\left(y_t\;|\; y_{1:t-1}\right)}{\pi_\text{ref}\left(y_t\;|\; y_{1:t-1}\right)}$. We derive  the expected value of the sequence-level KL gradient in the following $\nabla\mathbb{KL}_\text{seq} \left(\pi,\pi_\text{ref}\right)$.
\begin{align*}
      &=_{(a)} \nabla \mathbb{E}_{y\sim \pi}\left[\sum_{t=1}^T \log \rho_t\right] \\
    &=_{(b)} \mathbb{E}_{y\sim \pi}\left[\sum_{t=1}^T \nabla \log \pi\left(y_t\;|\; y_{1:t-1}\right) \sum_{s=1}^T \log \rho_s\right] \\
    &=_{(c)} \mathbb{E}_{y\sim \pi}\left[\sum_{t=1}^T \nabla \log \pi\left(y_t\;|\; y_{1:t-1}\right) \sum_{s=t}^T \log \rho_s\right] \\
    &=_{(d)}\underbrace{\mathbb{E}_{y\sim \pi}\left[\sum_{t=1}^T \nabla \log \pi\left(y_t\;|\; y_{1:t-1}\right) \log \rho_t\right]}_{\text{part i}} \\
    &+ \underbrace{\mathbb{E}_{y\sim \pi}\left[\sum_{t=1}^T \nabla \log \pi\left(y_t\;|\; y_{1:t-1}\right) \sum_{s=t+1}^T \log \rho_s\right]}_{\text{part ii}}.
\end{align*}
Here, (a) is by definition; (b) is from applying REINFORCE \citep{williams1992simple} to the expectation, while the path-wise gradient evaluates to zero for the KL divergence; (c) is is based on the fact that since $y_t$ does not impact past $\rho_s$ with $s<t$, and the related terms evaluate to zero. Finally, (d) breaks the summation into two parts. We see that the first term (part i) rewrite as
\begin{align*}
    \mathbb{E}_{y\sim \pi}\left[\sum_{t=1}^T \nabla\mathbb{KL}_t\right],
\end{align*}
i.e., the sum of token-level KL gradient across steps. When the token-level gradient is unbiased $\mathbb{E}\left[\nabla \widehat{\mathbb{KL}}_t\right]=\mathbb{E}\left[\nabla \mathbb{KL}_t\right]$ (i.e., the vanilla gradient estimate), we have that Eqn~\eqref{eq:partial} is an unbiased estimate to part i gradient.

\section{Details on LLM experiments}\label{appendix:llm}

Throughout, we experiment with the open sourced Llama 3.1 instruction-tuned models \citep{grattafiori2024llama} of size 8B and 70B. The reward maximization and on-policy distillation experiments are both carried out on the $7500$ MATH training prompts \citep{hendrycks2021measuring}. We use $n=4$ samples per gradient estimate.

\end{document}